\documentclass{article}

\usepackage{basic}

\usepackage[utf8]{inputenc} % allow utf-8 input
\usepackage[T1]{fontenc}    % use 8-bit T1 fonts
\usepackage{url}            % simple URL typesetting
\usepackage{booktabs}       % professional-quality tables
\usepackage{amsfonts}       % blackboard math symbols
\usepackage{nicefrac}       % compact symbols for 1/2, etc.
\usepackage{microtype}      % microtypography
\usepackage{lineno,hyperref}
\usepackage{graphicx}
\usepackage{multirow}
\usepackage{amsmath}
\usepackage{tabularx}
\newcolumntype{L}{>{\raggedright\arraybackslash}X}
\newcolumntype{C}{>{\centering\arraybackslash}X}
\usepackage{setspace}
\usepackage[none]{hyphenat}
\title{Positional Attention-based Frame Identification with BERT: A Deep Learning Approach to Target Disambiguation and Semantic Frame Selection}

\author{
  Sang-Sang Tan\thanks{Corresponding author} \\
  Wee Kim Wee School of Communication and Information\\ Nanyang Technological University\\
  31 Nanyang Link, Singapore 637718\\
  \texttt{tans0348@ntu.edu.sg} \\
   \And
 Jin-Cheon Na \\
  Wee Kim Wee School of Communication and Information\\ Nanyang Technological University\\
  31 Nanyang Link, Singapore 637718\\
  \texttt{tjcna@ntu.edu.sg} \\
}

\begin{document}
\sloppy

\maketitle

\begin{abstract}
Semantic parsing is the task of transforming sentences from natural language into formal representations of predicate-argument structures. Under this research area, frame-semantic parsing has attracted much interest. This parsing approach leverages the lexical information defined in FrameNet to associate marked predicates or targets with semantic frames, thereby assigning semantic roles to sentence components based on pre-specified frame elements in FrameNet. In this paper, a deep neural network architecture known as Positional Attention-based Frame Identification with BERT (PAFIBERT) is presented as a solution to the frame identification subtask in frame-semantic parsing. Although the importance of this subtask is well-established, prior research has yet to find a robust solution that works satisfactorily for both in-domain and out-of-domain data. This study thus set out to improve frame identification in light of recent advancements of language modeling and transfer learning in natural language processing. The proposed method is partially empowered by BERT, a pre-trained language model that excels at capturing contextual information in texts. By combining the language representation power of BERT with a position-based attention mechanism, PAFIBERT is able to attend to target-specific contexts in sentences for disambiguating targets and associating them with the most suitable semantic frames. Under various experimental settings, PAFIBERT outperformed existing solutions by a significant margin, achieving new state-of-the-art results for both in-domain and out-of-domain benchmark test sets.
\end{abstract}

% keywords can be removed

\section{Introduction}

Semantic parsing is essential for many knowledge-based applications that require semantic reasoning and interpretation of natural language. Such applications include question answering \cite{narayanan2004question, shen2007using}, information extraction \cite{christensen2010semantic, surdeanu2003using}, and knowledge graph construction \cite{gangemi2016identifying, alam2017event}, among many others. Given a sentence, the general goal of a semantic parser is to (1) identify predicates in the sentence, (2) detect sentence components that form the semantic arguments related to the predicates, and (3) determine the roles (e.g., Agent, Patient, Temporal, Manner, Cause, and so forth) of the semantic arguments. Through the detection of predicates and their semantic arguments, semantic parsing allows sentences to be computationally interpreted to answer the question ``Who did What to Whom, and How, When and Where?'' \cite[p.1]{palmer2010semantic}.

Recent years have seen increasingly rapid advances in semantic parsing due to the continuing efforts devoted to creating and maintaining semantically-annotated lexical resources \cite{baker1998berkeley, palmer2005proposition}. These resources come with manually-labeled text corpora, which are often used as development data to build statistical models for automated semantic parsing. In this study, we focus on the frame identification subtask in frame-semantic parsing. This parsing approach makes use of the lexical information defined in FrameNet \cite{baker1998berkeley} to generate semantic representations from natural language. As a knowledge base stemmed from the theory of Frame Semantics \cite{fillmore1976frame}, FrameNet is centralized around the idea that semantic frames are the conceptual categories and contexts that become the basis for meaning representation. Figure \ref{fig:frame-semantic-parsing} shows an example of frame-semantic parsing and a simplified schematic representation of the [Desirability] frame retrieved from FrameNet 1.7. The majority of work in frame-semantic parsing has concentrated on the following subtasks: 

\begin{figure}
  \includegraphics[width=\linewidth]{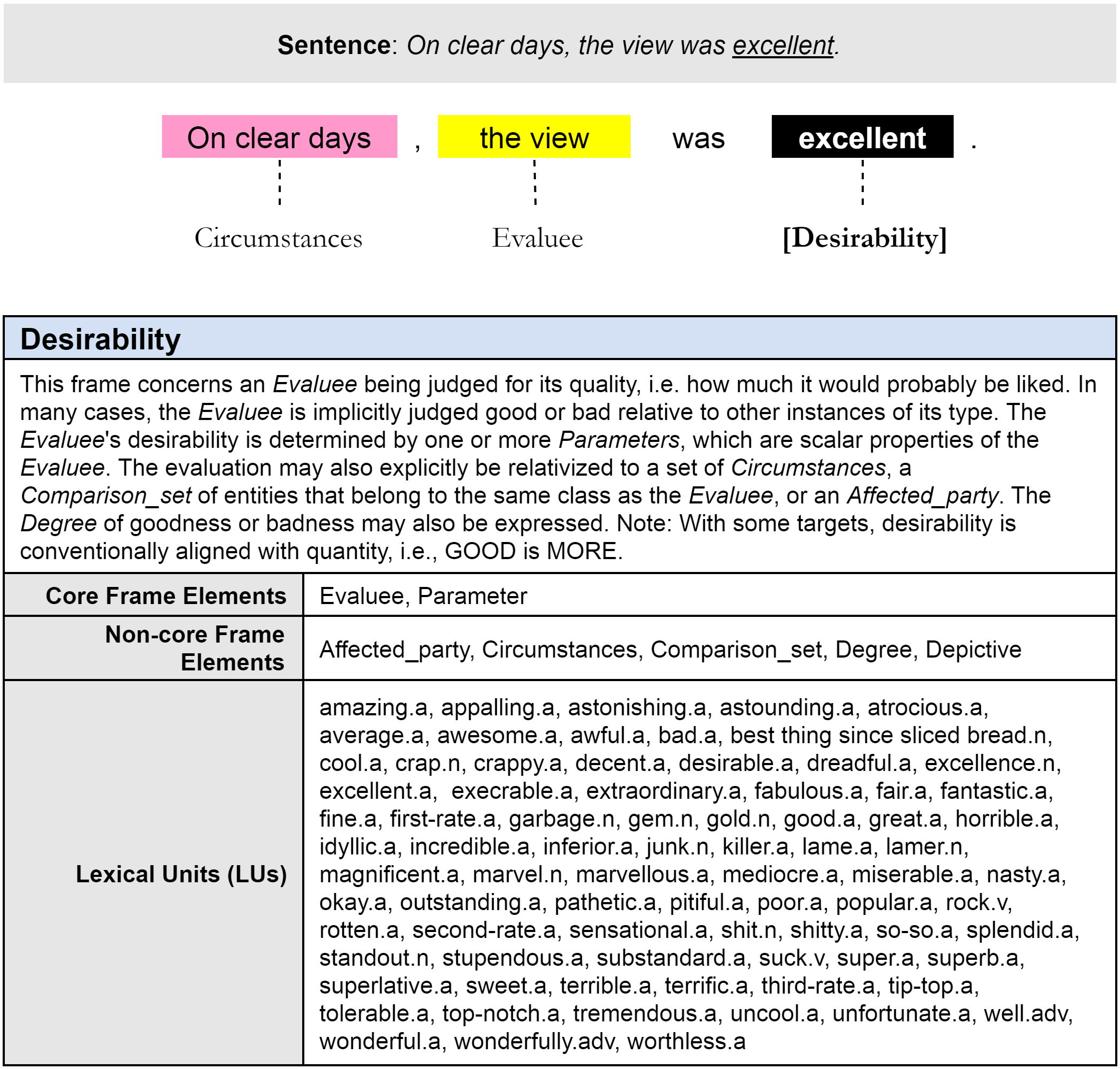}
  \caption{An example of frame-semantic parsing (retrieved from FrameNet 1.7). The parsing process involves two steps: first, the frame identification step identifies [Desirability] as the most likely frame for `excellent'; second, the semantic role assignment step determines the roles (i.e., Circumstances and Evaluee) for the sentence components based on the identified frame.}
  \label{fig:frame-semantic-parsing}
\end{figure}

\begin{itemize}
\item Frame identification. Given a sentence, this task identifies the most likely semantic frame for a marked lexical predicate (also known as target), which can be any word or phrase in the sentence. This step makes use of the association relations between the frames and lexical units (LUs) defined in FrameNet.
\item Semantic role assignment. Given the target and its identified frame, the second subtask aims to associate sentence components with semantic roles, which are specified as frame elements in FrameNet.
\end{itemize}

Frame identification plays a vital role in frame-semantic parsing because recognizing frames correctly is essential for the subsequent step of semantic role assignment. Specifically, the frame identification step reduces the search scope for the most suitable semantic roles from hundreds of roles in FrameNet to at most 30 potential roles per frame \cite{hartmann2017out}. The study conducted by \cite{hartmann2017out} showed that frame identification errors could cause drastic degradation in the accuracy of role assignment.

Motivated by the need for a high-quality and robust frame identification solution, this study presents a new method called Positional Attention-based Frame Identification with BERT (PAFIBERT). The proposed method employs a deep learning neural network architecture that combines a position-based attention mechanism and a state-of-the-art pre-trained language model known as Bidirectional Encoder Representations from Transformers (BERT) \cite{devlin2018bert}. As will be discussed in more detail in section \ref{sec:method}, the position-based attention mechanism is essential for the frame identification problem, in which multiple occurrences of the same target (i.e., the same word or phrase) in a sentence may refer to different frames. Incorporating this attention mechanism thus allows PAFIBERT to distinguish different instances of the target based on their positions and select the most suitable frame for each instance by attending to its surrounding context.

This work is inspired by the recent promising achievements of pre-trained language models in a wide range of natural language processing (NLP) tasks \cite{devlin2018bert, howard2018universal, radford2018improving}. These language models are pre-trained via unsupervised learning, for which an enormous amount of textual data can be obtained easily. The pre-training stage allows the models to capture basic linguistic information such as word contexts and inter-sentence relations \cite{devlin2018bert}. During pre-training, the language models learn weights that can be fine-tuned later for downstream NLP tasks, thus eliminating the need to learn the weights from scratch in those tasks, for which labeled data are usually required but hard to come by.

The frame identification task can be regarded as a conceptualization process that maps a target to a more abstract concept (i.e., a semantic frame). Because of homonymy and polysemy, an ambiguous target may refer to multiple frames. For instance, `beat' may be associated with either the [Cause\_Harm] frame or the [Beat\_Opponent] frame, depending on the context in which it is used. In this regard, frame identification is analogous to a word sense disambiguation task that disambiguates a target based on its sentential context. Therefore, frame identification may benefit from the deep, bidirectional architecture of BERT, which is designed to capture contextual information in texts from both left-to-right and right-to-left directions. The contributions of this study are as follows:

\begin{itemize}
\item This study presents a novel neural network architecture called PAFIBERT that combines a position-based attention mechanism with BERT to solve the frame identification problem. By providing a solution for frame identification, this study makes a solid contribution to the field of semantic parsing and its potential application areas.
\item PAFIBERT produced new state-of-the-art results on two benchmark datasets: the in-domain Das's test set \cite{das2011semi} and the out-of-domain YAGS test set \cite{hartmann2017out}. The former consists of data held out from FrameNet's text corpora, whereas the latter contains data collected from other sources. To date, most studies have only reported test results for the former. However, as demonstrated by \cite{hartmann2017out}, there exists a disparity between the results produced from the in-domain and out-of-domain evaluations. Their findings have suggested that the out-of-domain evaluation is crucial in frame identification as its results serve as an indicator of model generalizability in real-world applications. The present study not only included the YAGS test set but also examined PAFIBERT qualitatively on other out-of-domain sample sentences that contain interesting cases of unseen targets. The results obtained in these quantitative and qualitative out-of-domain evaluations demonstrated the robustness and generalizability of our method.
\item In addition to obtaining promising results in frame identification, the proposed neural network design also provides insights into the adaptation of BERT \cite{devlin2018bert}, not only for this specific problem, but also for other similar NLP problems that rely on sentential contexts to make target-specific predictions. Examples of such problems include word sense disambiguation, aspect-based sentiment analysis, and named entity recognition, among others. The findings and achievements in this study provide convincing empirical proof that adds to the rapidly expanding body of research on language modeling and transfer learning in NLP \cite{devlin2018bert, howard2018universal, radford2018improving}.
\end{itemize}

\section{Related Work}

\subsection{Frame Identification}

Frame-semantic parsing has started to attract more interest after the introduction of the SemEval 2007 shared task on frame-semantic structure extraction \cite{baker2007semeval}. The best system in the shared task was presented by \cite{johansson2007lth}. For the frame identification subtask, the researchers employed Support Vector Machines (SVMs) to learn a set of classifiers using the following features: lemmatized targets, word forms of the targets, features generated from dependency parsing such as dependents and parents of the targets, and so forth. This system was then outperformed by an earlier version of SEMAFOR \cite{das2010semafor}, a frame-semantic parser that used a conditional log-linear model for frame identification. A key challenge in frame identification is dealing with unseen targets, i.e., the targets that do not exist in the FrameNet lexicon and the hand-labeled text corpora released along with FrameNet. While the best system in the SemEval 2007 shared task tried to handle unseen targets by expanding the LU coverage of FrameNet using WordNet synsets \cite{johansson2007lth}, the approach proposed by \cite{das2010semafor} used a latent variable to incorporate lexical-semantic relations into their model. These lexical-semantic features allowed the model to make predictions based on the relations between unseen targets and FrameNet's LUs. Further improvement was also suggested by \cite{das2011semi} to address the problem of unseen targets. They adopted a semi-supervised learning approach to build a graph from FrameNet's text corpora and a large amount of unlabeled data. The graph vertices corresponded to targets, whereas the edges represented frame similarity and distributional similarity. For the known targets (a subset of the graph vertices), their weights were initialized based on the frame distribution in FrameNet's text corpora. Through a label propagation process, the frames (i.e., labels) were transferred from the known targets to the unlabeled vertices of the graph. The constructed graph was then used to suggest candidate frames for unseen targets.

Earlier work on frame identification mainly focused on feature-based approaches, which relied on hand-engineered features. Since the popularization of distributed representations \cite{mikolov2013efficient}, more recent work in frame identification has gradually shifted toward the use of word embeddings or low-dimensional vector representations. The word embedding approach not only reduces the need for expensive feature engineering of semantic structures but also leads to improved accuracy due to better generalization \cite{fitzgerald2015semantic}. One of the earliest studies that leveraged distributed representations for frame identification was carried out by \cite{hermann2014semantic}. Their technique extracted the syntactic contexts of targets from data and generated an initial high-dimensional vector space using word embeddings. The WSABIE algorithm \cite{weston2011wsabie} was then used to find a transformation function that mapped the initial high-dimensional vectors to low-dimensional representations. Embeddings were also learned for frame labels in the low-dimensional space such that for each target, the proximity between the target and its matching frame (i.e., gold frame) exceeded that between the target and the competing frames. At inference time, after a target and its context were projected using the learned function to the low dimensional space, the nearest frame label was chosen as the prediction. The accuracy achieved by this technique was highly promising, surpassing the results reported in previous studies by a large margin. However, for the generation of the initial high-dimensional vectors, this technique still involved a great deal of feature engineering to extract features via syntactic parsing. In the same vein, a study conducted by \cite{hartmann2017out} experimented with pre-trained word embeddings to represent targets and their contexts as inputs for learning two classification models: a two-layer neural network and a WSABIE model as proposed by \cite{hermann2014semantic}. Although their models did not outperform that of \cite{hermann2014semantic} for Das's benchmark test set \cite{das2011semi}, the results they obtained on out-of-domain data seemed to suggest that their models generalized better in other domains. 

Following the success of deep learning in many NLP tasks, researchers in frame-semantic parsing have also explored various types of neural networks for frame identification. The method proposed by \cite{swayamdipta2017frame} combined pre-trained word embeddings, learned word embeddings, and part-of-speech (POS) embeddings for frame identification using a long short-term memory (LSTM) neural network and achieved an accuracy similar to that of \cite{hartmann2017out}. Another study \cite{yang2017joint} examined the effectiveness of a multi-layer neural network for frame identification and found that the overall model performance was competitive, and their method outperformed other models in predicting frames for ambiguous LUs. 

In a recent study, \cite{peng2018learning} proposed a frame identification solution that employed a joint decoding approach. This approach used disjoint data to improve model performance in multi-task scenarios by allowing information sharing across tasks to benefit all the tasks. For the frame-semantic parsing problem, which consists of the frame identification and semantic role assignment subtasks, the joint decoding approach could learn model parameters jointly for both tasks by optimizing a multi-task objective. Their solution showed significant improvements over previous work, producing state-of-the-art accuracies of 90\% and 89.1\% for frame identification on FrameNet 1.5 and FrameNet 1.7 respectively.

Although extensive research has been carried out in this area, as pointed out by \cite{hartmann2017out}, most studies have only focused on increasing model accuracies for Das's benchmark test set \cite{das2011semi}. Without investigating rigorously the performance of frame identification models on other datasets that may contain more cases of unseen targets, domain-specific words, and jargon, it is hard to tell how well these models would perform in real-world NLP applications. The present study thus set out to develop a robust solution that not only outperforms other models on Das's test set but also works reliably on other data.

\subsection{Pre-trained Language Models}

There are various ways to convert textual data into the numerical representations that are processable by machine learning algorithms. Local representations in the form of sparse, binary vectors can be obtained easily through one-hot encoding, but they tend to suffer several problems such as the curse of dimensionality and lack of semantics \cite{bengio2000modeling, zhang2016deep}. To overcome the limitations of local representations, research in distributional semantics \cite{mikolov2013efficient, bruni2014multimodal, mikolov2013distributed, mitchell2010composition} aims to produce numerical representations for linguistic items based on their distributions in textual data. This line of research has led to the popularization of word embedding, which has become a standard representation technique in NLP. In contrast to the local representations generated from one-hot encoding, distributed representations use low-dimensional vectors of real numbers to capture the semantic and syntactic relations between linguistic items \cite{mikolov2013efficient, mikolov2013distributed}. 

Generating high-quality word embeddings efficiently is of significant interest to the research community. In particular, the production and optimization of pre-trained word embeddings have been an active research area (e.g., \cite{mikolov2017advances, pennington2014glove}). Pre-trained word embeddings can be obtained via unsupervised learning on unlabeled data. Since the pre-training process is task-agnostic and domain-independent, the resulted word vectors are applicable to a wide range of downstream tasks. Apparently, pre-trained word vectors such as those produced by GloVe \cite{pennington2014glove} and Word2Vec \cite{mikolov2013efficient} have brought NLP a long way. However, this type of pre-training approach is considered shallow because word embeddings usually serve as input features to initialize the first layer of neural network models, but the rest of the models are trained from scratch using task-specific data \cite{howard2018universal}. There have been some attempts to incorporate pre-trained embeddings into other layers of neural networks, but since the embeddings are still treated as fixed weights in those models, their effectiveness and usefulness are somewhat limited \cite{howard2018universal}.

Outside of NLP, transfer learning in computer vision research has shown revolutionary improvements in image classification tasks \cite{he2016deep, krizhevsky2012imagenet, simonyan2014very}. It is thus not surprising that transfer learning has also attracted tremendous research interest among the NLP community in recent years. Transfer learning usually involves fine-tuning a pre-trained model, i.e., a model that was trained in an unsupervised setting on a massive unlabeled dataset. Instead of learning from scratch, the learning process for a downstream task can then benefit from the weights of the pre-trained models. In other words, the process only needs to adjust the weights to adapt to the specific task, usually with supervised learning on a relatively smaller set of labeled data. Compared to using pre-trained word embeddings, employing a pre-trained language modeling approach allows all model weights---including weights in the pre-trained layers and the task-specific layers---to be jointly updated for a downstream NLP task. Recent advancements in NLP have acquired convincing empirical proof that supports the shift of paradigm from shallow to deep pre-training \cite{devlin2018bert, howard2018universal, radford2018improving}. Highly promising results have been achieved using pre-trained language models like Universal Language Model Fine-tuning (ULMFiT) \cite{howard2018universal}, OpenAI's Generative Pre-trained Transformer \cite{radford2018improving}, and Google's Bidirectional Encoder Representations from Transformers (BERT) \cite{devlin2018bert} on a wide range of NLP tasks.

BERT was introduced by \cite{devlin2018bert} to overcome some limitations in the implementation of language models. They argued that the unidirectional architecture of language models like OpenAI's GTP \cite{radford2018improving} could be suboptimal. BERT is a deep bidirectional model trained using two novel pre-training tasks that allowed the language model to capture (1) the contextual information of tokens from both left-to-right and right-to-left directions and (2) the relations between sentences. Although the concept underlying the development of BERT is simple, it has been empirically proven as more potent than other pre-trained language models, rewriting the records of state-of-the-art results on benchmark datasets in 11 NLP tasks.

\section{The Proposed Method} \label{sec:method}

We propose an attention-based neural network architecture called Positional Attention-based Frame Identification with BERT (PAFIBERT) as a solution to the frame identification problem. In NLP, the use of attention mechanism has its root in machine translation \cite{bahdanau2014neural}. It has since become an essential component of neural network models for many NLP problems like text summarization (e.g., \cite{nallapati2016abstractive, rush2015neural}), question answering (e.g., \cite{xiong2016dynamic, yang2016anmm}), and aspect-based sentiment analysis (e.g., \cite{ma2017interactive, wang2016attention}). In general, the primary focus of attention mechanisms in deep learning is to compute alignment vectors consisting of weights that represent the importance of other elements in neural networks. Our method makes use of targets' positions to attend to the sentence constituents that are useful for disambiguating the targets. In the following subsections, we define the frame identification problem in detail and describe our solution to the problem.

\subsection{Problem Definition} \label{sec:problem-definition}

\begin{figure}
  \includegraphics[width=\linewidth]{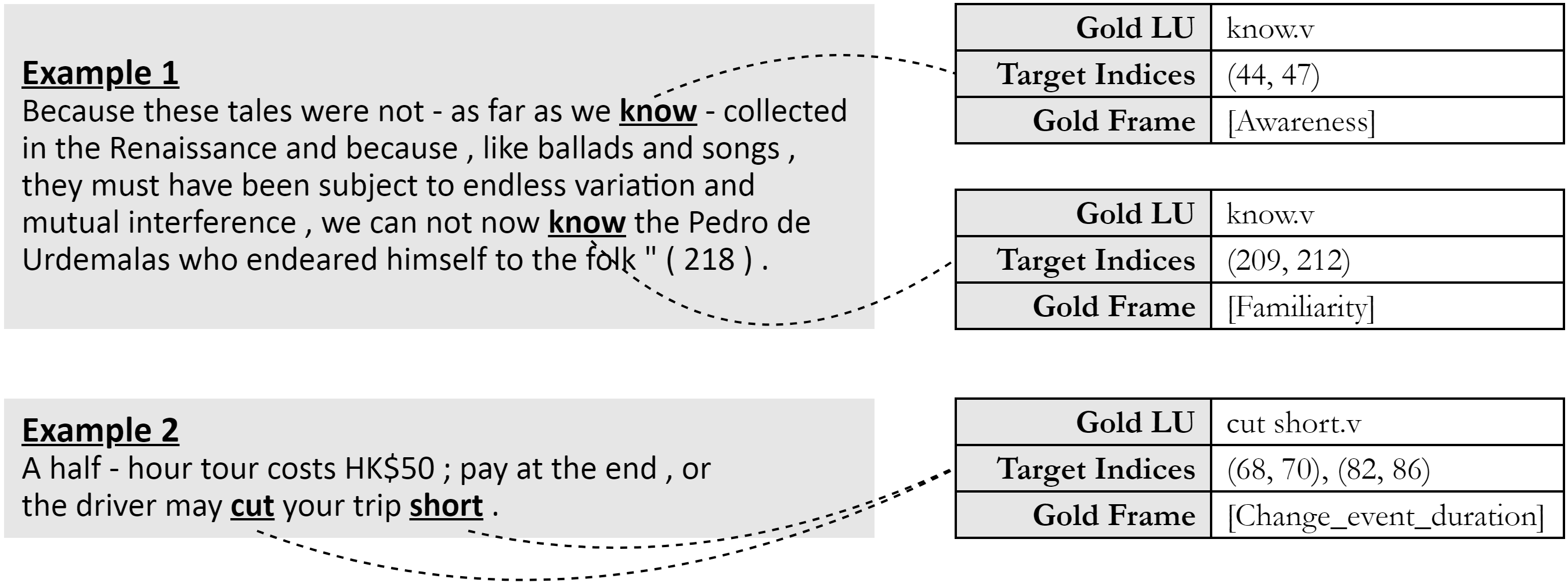}
  \caption{Examples of FrameNet annotations}
  \label{fig:framenet-annotation}
\end{figure}

Several examples of FrameNet annotations for frame identification are given in Figure \ref{fig:framenet-annotation}. These examples are revealing in several ways: first, they show that targets are marked by indices that indicate their positions in sentences; second, a target may consist of non-contiguous stretches of words (as in Example 2); third, as demonstrated by Example 1, multiple instances of the same target may occur at different positions in a sentence and be associated with different frames.

Given a sentence $s$ and a set of indices $x = \{(a_{1}, b_{1}), (a_{2}, b_{2}), \ldots\}$ that indicates the positions of a target, the goal of frame identification is to determine the most suitable frame for the target based on its usage in $s$.

\subsection{Positional Attention-based Frame Identification with BERT (PAFIBERT)}

The PAFIBERT method proposed in this study is built on top of BERT. The model architecture used by BERT is a multi-layer transformer encoder, implemented based on the original transformer architecture proposed by \cite{vaswani2017attention}. Two BERT models of different sizes are available: $BERT_{BASE}$ and $BERT_{LARGE}$. This study used $BERT_{BASE}$, which consists of 12 transformer blocks, 768 hidden dimensions, and 12 self-attention heads. Input sequences of BERT are constructed by combining token embeddings, segment embeddings, and position embeddings. Special tokens like $[CLS]$ and $[SEP]$ are used to indicate the beginning of sequences or separate non-consecutive sequences in multi-sequence tasks. BERT adopts the WordPiece technique \cite{wu2016google} for token representation, and word pieces or subwords are preceded by $\#\#$.

BERT is designed to handle various types of downstream NLP tasks, including those requiring sentence pairs instead of single sentences as inputs. As demonstrated by \cite{devlin2018bert}, BERT could be adopted easily for numerous NLP tasks by simply adding an output layer to the model. Several task-specific models were presented in their work for common NLP tasks like sentiment analysis, sentence similarity analysis, and textual entailment. However, as described in subsection \ref{sec:problem-definition} of this paper, for the frame identification problem, an ambiguous target with multiple occurrences in a given sentence may be associated with different frames, depending on its context or usage in the sentence. None of the task-specific models introduced by \cite{devlin2018bert} provides a valid solution to this problem. For example, a model that uses sentence-target pairs as inputs would generate the same input representation for all occurrences of the target in the same sentence. This limitation prevents the model from disambiguating different instances of the target. As a remedy to this problem, PAFIBERT leverages targets' positions for position-based attention, thus allowing the generation of target-specific representations for frame identification.

\begin{figure}
  \includegraphics[width=\linewidth]{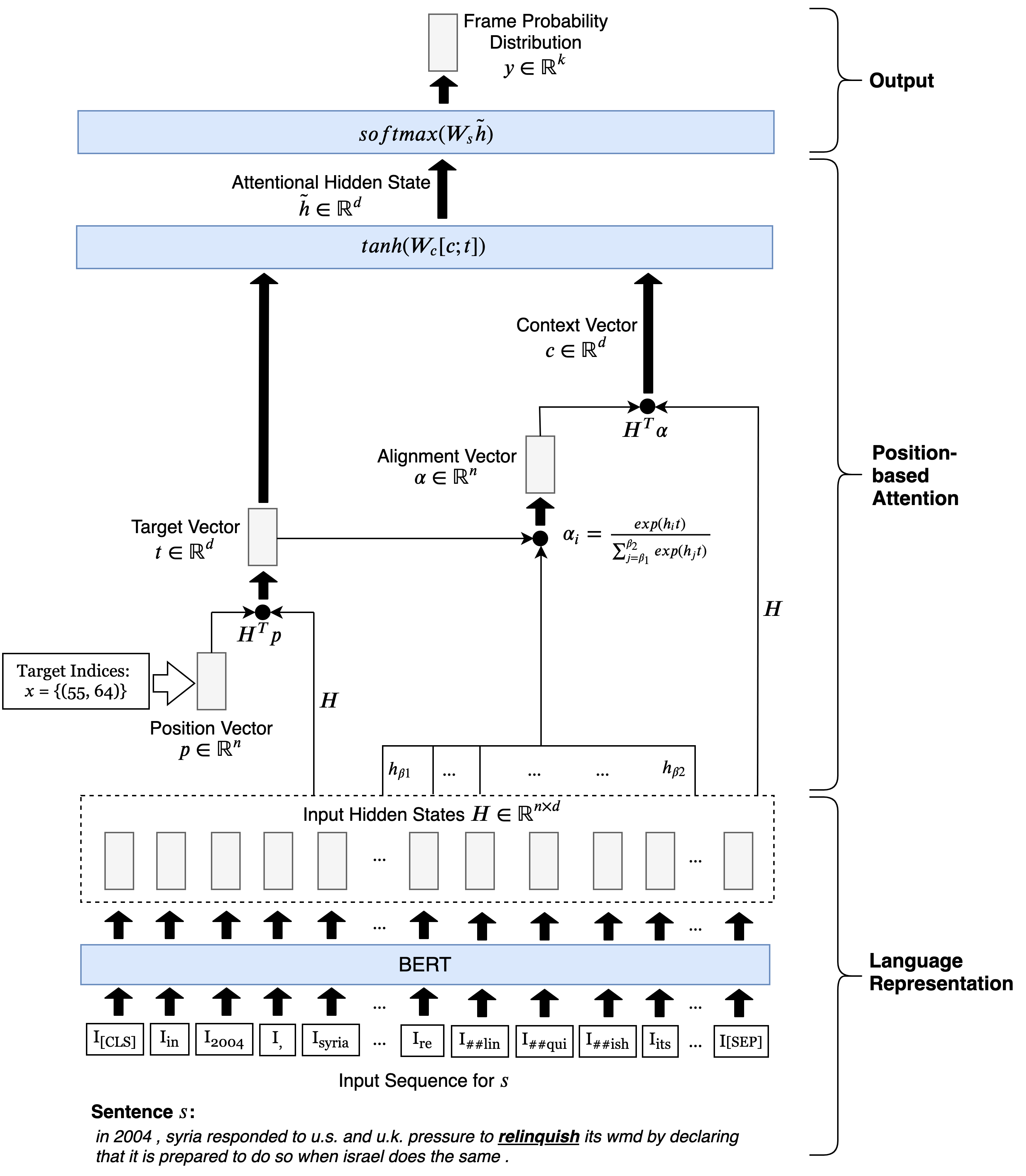}
  \caption{PAFIBERT without frame filtering}
  \label{fig:pafibert-nofil}
\end{figure}

\begin{figure}
  \includegraphics[width=\linewidth]{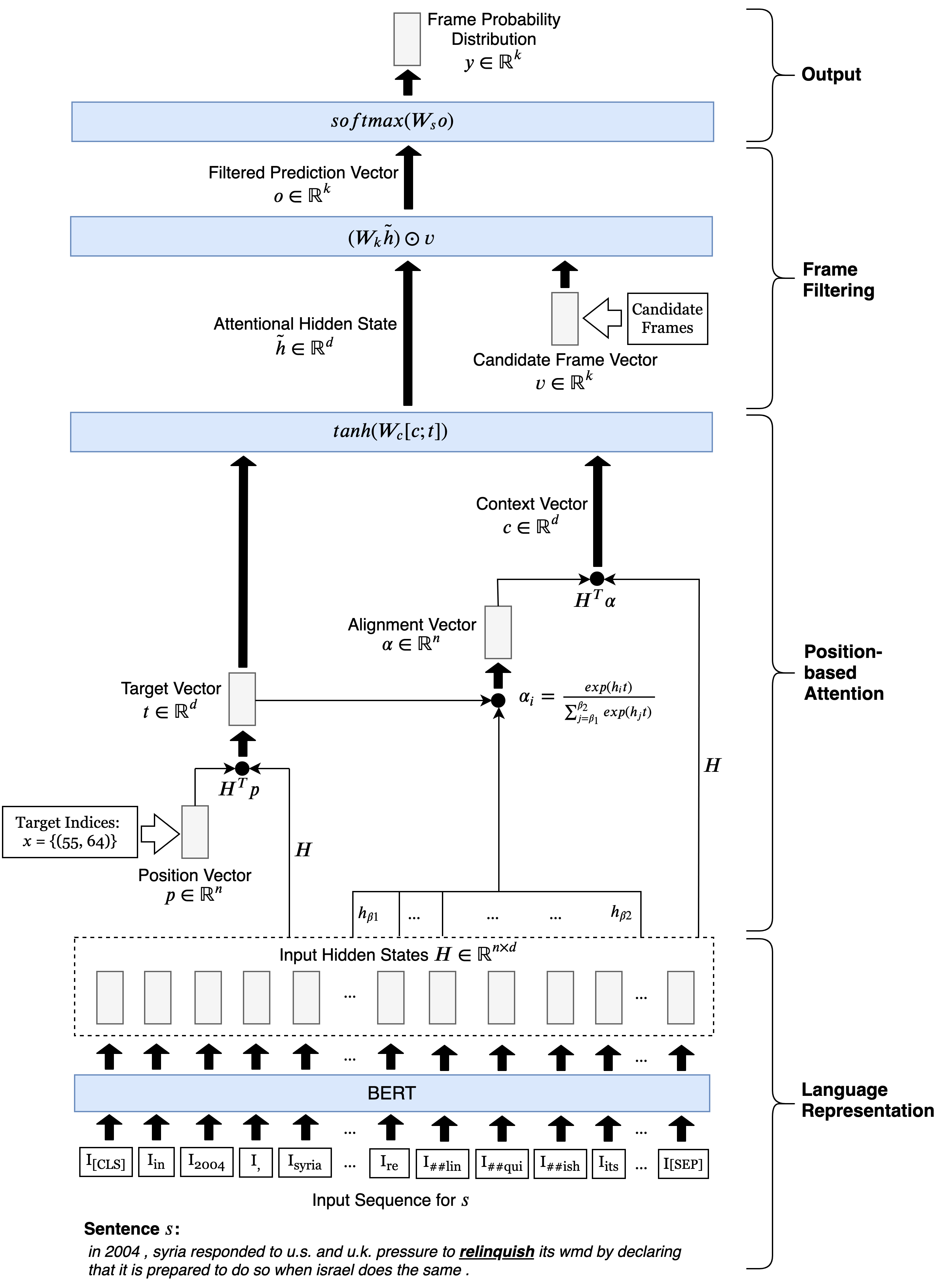}
  \caption{PAFIBERT with frame filtering}
  \label{fig:pafibert-fil}
\end{figure}

Two variants of PAFIBERT are presented in this study: without frame filtering and with frame filtering. The two neural network designs are depicted in Figure \ref{fig:pafibert-nofil} and \ref{fig:pafibert-fil}. The key difference between these two variants will be discussed in the next subsection, but in general, PAFIBERT needs two inputs: sentence $s$ and a target indicated by its position indices $x$. The input sequence generated from $s$ is fed through the transformer layers of BERT to produce $H$, an $(n \times d)$ matrix of hidden states where $n$ is the sequence length and $d$ is the hidden dimension. The attention mechanism transforms $H$ into a pooled attentional hidden state $\tilde{h}$, which is a $d$-dimensional vector that captures the information about the input sentence and the target. Using the default configuration in $BERT_{BASE}$, the hidden dimension is set to 768. As suggested by \cite{luong-etal-2015-effective}, the attentional hidden state can be computed as
\begin{equation}
\label{eq:1}
\tilde{h} = tanh(W_{c}[c;t])
\end{equation}
where the context vector $c$ and target vector $t$ are combined via concatenation and passed through a neural network layer. $W_{c}$ is the weight matrix of the layer, and the context vector $c$ is a $d$-dimensional vector derived as follows:
\begin{equation}
\label{eq:2}
c = H^T\alpha
\end{equation}

The context vector $c$ captures the relations between the hidden states and the $n$-dimensional alignment vector $\alpha$, which defines how much each hidden state contributes to the prediction of an output. Various alternatives exist for computing the alignment vector $\alpha$ in the above equation. PAFIBERT adopts a local attention mechanism \cite{luong-etal-2015-effective} that restricts the attention to only a subset of the input hidden states. Compared to a global attention mechanism that computes the alignment scores for all hidden states of an input sequence, a local attention mechanism is less expensive in terms of computational costs. The intuition of using a local attention mechanism for the frame identification problem is that, given a target in a lengthy sentence, frame identification models may not need the whole sentence to predict the most likely frame for the target. Instead, the contextual words adjacent to the target are likely to be sufficient for this task. Hence, PAFIBERT applies a fixed window ($w$) to limit the scope of attention to only $w$ hidden states before and after the target positions. In other words, it only computes the alignment scores for the hidden states that fall within the range of $[\beta_{1}, \beta_{2}]$ where
\begin{align}
\beta_{1} &= 
  \begin{cases}
    p_{start}-w & \text{if } (p_{start}-w)> 1,\\
    1 & \text{if } (p_{start}-w)\leq 1,
  \end{cases}\\
\beta_{2} &=
  \begin{cases}
    p_{end}+w & \text{if } (p_{end}+w) < n,\\
    n & \text{if } (p_{end}+w)\geq n.
  \end{cases}
\end{align}

In the above formulas, $p_{start}$ and $p_{end}$ denote the smallest and largest target positions ($p_{start}$ = $p_{end}$ for single-word targets). By observing the typical scope needed to infer the correct frames for the hand-labeled annotations in FrameNet's text corpora, we selected the size of the fixed window as $w = 10$ in this study. For each hidden state at position $i$ (i.e., for the $i^{th}$ input token) within the range of $[\beta_{1}, \beta_{2}]$, its alignment score $(\alpha_{i})$ is calculated as follows:
\begin{equation}
\label{eq:5}
\alpha_{i} = \frac{exp(h_{i}t)}{\sum_{j=\beta_{1}}^{\beta_{2}} exp(h_{j}t)}
\end{equation}

\begin{figure}
  \includegraphics[width=\linewidth]{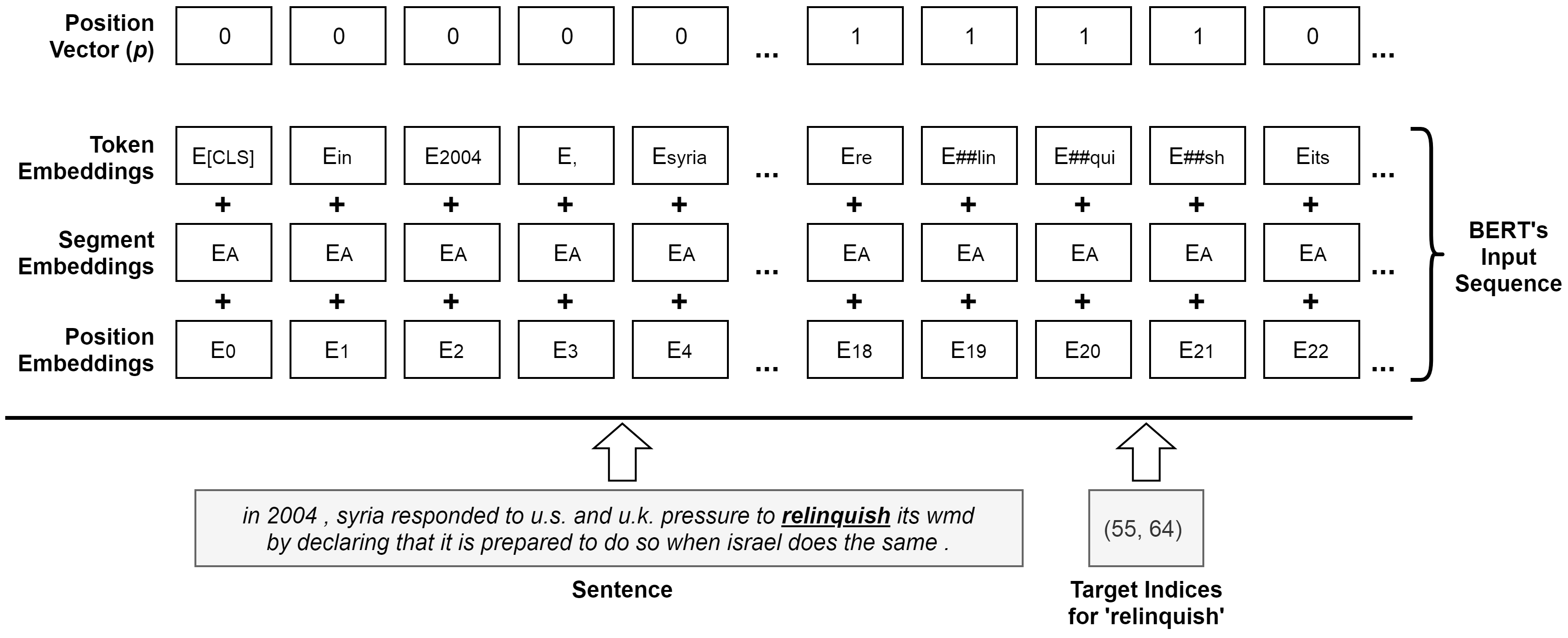}
  \caption{An example of the position vector $p$}
  \label{fig:position-vector}
\end{figure}

Note that Equation \ref{eq:5} describes a softmax function such that $\alpha$ represents a probability distribution that defines the importance of each hidden state. For the hidden states that fall outside the fixed window, their alignment scores are set to zeros. The derivation of $t$---the target vector---in Equation \ref{eq:1} and \ref{eq:5} is very much problem-specific and varies across different attention-based neural networks. In our method, $t$ is extracted from the hidden states that correspond to the positions of a specified target. The target positions are represented using an $n$-dimensional position vector ($p$), which consists of binary values indicating whether a particular position is a target position (value 1) or a non-target position (value 0). Since BERT uses the WordPiece technique \cite{wu2016google} for text tokenization, the position vector $p$ is derived such that it is compatible with BERT's input tokens. As depicted in Figure \ref{fig:position-vector}, if a target is split by BERT's tokenizer into word pieces, the position vector marks the word pieces as part of the target. Based on the target's positions, the $d$-dimensional target vector is then derived as follows from the hidden states in $H$:
\begin{equation}
\label{eq:6}
t=H^Tp
\end{equation}

\subsection{Frame Filtering}

Most frame identification studies have adopted the standard experimental setup in which FrameNet is used to specify the candidate frames given a target, and the most suitable frame for the target is then selected from the candidate frames \cite{hartmann2017out}. Under this setup, if the target is unambiguous---it has only one candidate frame---the frame is selected directly. On the other hand, if the target is ambiguous, this frame filtering step allows frame identification models to limit their search to the candidate frames instead of considering every frame in FrameNet. This setup, however, makes strong assumptions about the comprehensiveness of FrameNet: It assumes that FrameNet specifies every possible sense of the target and that the list of LUs under each frame is exhaustive. These assumptions may be acceptable when applied to the development data released by FrameNet, but they certainly do not hold for real-world data. Therefore, following the suggestion of \cite{hartmann2017out}, we experimented with two neural network designs: with frame filtering and without frame filtering.

In the no-filtering model, the attentional hidden state ($\tilde{h}$) is simply fed through a softmax layer to generate a $k$-dimensional vector $y$ using Equation \ref{eq:7} below, where $k$ is the number of frames in FrameNet. Vector $y$ denotes the frame probability distribution given sentence $s$ and a target marked by position indices $x$. $W_{s}$ is the weight matrix of the softmax layer.
\begin{equation}
\label{eq:7}
y=softmax(W_{s}\tilde{h})
\end{equation}

In the model with frame filtering, the filtering step is added before the softmax layer so that the probability values for the non-candidate frames are set to zeros, whereas the probability values for the candidate frames sum up to 1. This filtering step generates the filtered prediction vector $o$, which is a $k$-dimensional vector calculated using the formula below:
\begin{equation}
\label{eq:8}
o=(W_{k}\tilde{h}) \odot v
\end{equation}

In the above equation, $\tilde{h}$ is passed through a fully-connected layer consisting of $k$ nodes. $W_{k}$ is the weight matrix of the fully-connected layer. Element-wise multiplications are then performed between the output of the full-connected layer and vector $v$, which is a $k$-dimensional vector consisting of binary values that specify whether a particular frame is a candidate frame. For unseen targets, this step assumes that every frame in FrameNet is a potential candidate. Due to the element-wise multiplications, only the candidate frames will have non-zero probability values. Finally, the model produces the $k$-dimensional frame probability distribution as follows:
\begin{equation}
\label{eq:9}
y=softmax(W_{s}o)
\end{equation}

\subsection{Listing of Candidate Frames}

As shown in Figure \ref{fig:framenet-annotation}, FrameNet annotations associate each target with a gold LU and gold frame. Therefore, when implementing frame filtering in frame identification models, existing studies (e.g., \cite{hermann2014semantic, swayamdipta2017frame, peng2018learning}) have assumed the gold LUs are known. In other words, the models were allowed to use the gold LUs to retrieve candidate frames from FrameNet.

Apparently, this simplified setup does not reflect real-world deployment scenarios accurately. In a more realistic setup, the gold LUs should be unknown, and frame identification models should predict based on the given targets, which may occur in any word forms (e.g., singular or plural form, past tense or present tense, comparative or superlative form, and so forth). When applying frame filtering to PAFIBERT, we experimented with the following approaches:

\begin{itemize}
\item Frame filtering by LUs. This is the simplified approach described above, in which the gold LUs are used to retrieve candidate frames for frame filtering.
\item Frame filtering by targets. This approach assumes that the gold LUs are unknown. Additional pre-processing steps are thus needed to enumerate potential candidate frames for frame filtering. To this end, we utilized the Pattern toolkit (\url{https://www.clips.uantwerpen.be/pattern}) to generate derived forms such as comparatives, superlatives, conjugated verbs, and so forth for every LU in FrameNet. A lookup table was then created to link these inflections to potential candidate frames. During model training and testing, the given targets were used as lookup keys to retrieve candidate frames from the table. For instance, for the LU `know.v', the table construction step generated `know', `knows', `knowing', and `knew' as the derived forms. When one of these derived forms was marked as the target, the lookup table returned [Certainty], [Differentiation], [Awareness], and [Familiarity] as candidate frames.
\end{itemize}

\section{Experiments}

\subsection{Model Training and Hyperparameter Tuning}

The latest version of FrameNet is 1.7, but FrameNet 1.5 has been the version used almost exclusively in frame-semantic research. Compared to FrameNet 1.5, FrameNet 1.7 has a substantially larger number of frames and LUs, thus presenting a more challenging stage for frame identification. To facilitate the comparison of results with other studies, we conducted our experiments using both FrameNet versions. FrameNet provides the following types of hand-labeled data with each release:

\begin{itemize}
\item The exemplary data, which contain example sentences of the LUs. Each sentence is annotated for one target and LU.
\item The full-text data, which consist of sentences with multiple annotations pertaining to multiple targets and LUs.
\end{itemize}

Both the exemplary data and the full-text data were used in this study for model development. We used the same train/test split as in previous work, and 16 documents from the full-text data were set aside for hyperparameter tuning \cite{hermann2014semantic}. The hyperparameters were selected from the ranges of values suggested by \cite{devlin2018bert}. The ranges for the batch size and the learning rate are \{16, 32\} and \{2e-5, 3e-5, 5e-5\} respectively. However, we found that the range suggested for the number of epochs (i.e., \{3, 4\}) did not produce optimal models in our initial experiments. We thus increased the range to \{3, 4, 5, 6, 7, 8\} for the hyperparameter search. The maximum input sequence length ($n$) was set based on the length of the longest sentence in the training data. We chose $n = 260$ for FrameNet 1.5 and $n = 320$ for FrameNet 1.7. The selected hyperparameters are shown in Table \ref{tab:hyperparam}.

\begin{table}
\caption{The selected hyperparameters}
\label{tab:hyperparam}
\renewcommand\arraystretch{1.4}
\begin{tabularx}{\textwidth}{l L L L l L L L}
%%\hline\noalign{\smallskip}
\toprule
\multirow{2}{*}{} & 
\multicolumn{3}{c}{\textbf{FrameNet 1.5}} &&
\multicolumn{3}{c}{\textbf{FrameNet 1.7}} \\
%%\cline{2-4}\cline{6-8}\noalign{\smallskip}
\cmidrule{2-4}\cmidrule{6-8}
& \textbf{Filtered by LUs} & \textbf{Filtered by Targets} & \textbf{No Filter} && \textbf{Filtered by LUs} & \textbf{Filtered by Targets} & \textbf{No Filter} \\
%%\hline\noalign{\smallskip}
\midrule
Batch size & 32 & 32 & 32 && 16 & 16 & 16 \\
Learning rate & 3e-5 & 3e-5 & 3e-5 && 3e-5 & 3e-5 & 3e-5 \\
Epoch & 8 & 6 & 8 && 7 & 6 & 8 \\
%%\hline\noalign{\smallskip}
\bottomrule
\end{tabularx}
\end{table}

As described in the previous section, the following variants of PAFIBERT were developed in this study:
\begin{itemize}
\item No frame filtering
\item Frame filtering based on LUs
\item Frame filtering based on Targets
\end{itemize}

\subsection{Model Evaluation}

The above models were evaluated on two benchmark test sets: Das's test set \cite{das2011semi} and the YAGS test set \cite{hartmann2017out}. The former is a widely used test set randomly sampled by \cite{das2011semi} from FrameNet's full-text data. It is an in-domain test set because it shares the same data sources as the training data. In FrameNet 1.5, Das's test set contains 4,458 frame annotations from 23 documents. The number of frame annotations in these documents was increased to 6,728 with FrameNet 1.7. It is worth noting that some sentences in Das's test set also occur in FrameNet's exemplary data, but we removed the duplicated sentences from the training data to ensure that the test data were completely unseen during the model training stage. The YAGS test set was introduced by \cite{hartmann2017out} as an out-of-domain test set. This test set contains 1,415 sentences sampled from Yahoo! Answers, and the number of frame annotations is 3,091. Apparently, for a frame identification model to be usable for real-world problems, it is crucial to make sure the model remains robust for out-of-domain data. In order to evaluate our method more rigorously, in addition to the evaluation on the YAGS test set, we also examined PAFIBERT qualitatively on several sample sentences obtained from product reviews in a relatively under-explored domain, i.e., the beauty product domain. All tokens in the sample sentences, except certain stopwords, were passed as targets to our models. We included sentences that contain unseen targets and domain-specific terms to examine the capability of the models in handling these more complicated cases.

\section{Results and Discussion}

\subsection{Model Performance on Benchmark Test Sets}

Table \ref{tab:result-fil} and \ref{tab:result-nofil} compare the accuracies of our models to the results reported by several representative studies in frame-semantic parsing. Since the key challenge in frame identification lies in the ambiguous LUs, some studies have assessed these more challenging and interesting cases separately to provide a more useful performance measure. Likewise, we employed the same procedure to evaluate our models for (1) all targets and (2) the ambiguous targets.

As can be seen from the tables, PAFIBERT outperformed other models on both the in-domain Das's test set and the out-of-domain YAGS test set, yielding new state-of-the-art results for frame identification. As expected, the results obtained with LU-based frame filtering were slightly better than those produced using target-based frame filtering, suggesting that a realistic setup is crucial in measuring model performance more precisely. In the following discussion, we only refer to the target-based approach when the discussion involves PAFIBERT with frame filtering.

\begin{table}
\caption{Results with frame filtering}
\label{tab:result-fil}
\renewcommand\arraystretch{1.4}
\begin{tabularx}{\textwidth}{L c c c c c}
\toprule
\multirow{2}{*}{\textbf{Model}} &
\multicolumn{2}{c}{\textbf{FrameNet 1.5}} &&
\multicolumn{2}{c}{\textbf{FrameNet 1.7}} \\
\cmidrule{2-3}\cmidrule{5-6}
& \textbf{All} & \textbf{Ambiguous} && \textbf{All} & \textbf{Ambiguous} \\
\midrule
\multicolumn{6}{l}{\underline{\textit{Das's Test Set \cite{das2011semi}}}} \\
SEMAFOR \cite{das2014frame} & 83.60 & 69.19 && - & - \\
Hermann et al. \cite{hermann2014semantic} & 88.73 & 73.67 && - & - \\
Yang and Mitchell \cite{yang2017joint} & 88.20 & 75.70 && - & - \\
Hartmann et al. \cite{hartmann2017out} & 87.63 & 73.80 && - & - \\
Botschen et al. \cite{botschen-etal-2018-multimodal} & 88.82 & 75.28 && - & - \\
Peng et al. \cite{peng2018learning} & 90.00 & 78.00 && 89.10 & 77.50 \\
\textbf{PAFIBERT - Filtered by LUs} & \textbf{92.22} & \textbf{82.90} && \textbf{91.44} & \textbf{82.55} \\
\textbf{PAFIBERT - Filtered by Targets} & \textbf{91.39} & \textbf{82.80} && \textbf{90.15} & \textbf{81.92} \\
\midrule
\multicolumn{6}{l}{\underline{\textit{YAGS Test Set \cite{hartmann2017out}}}} \\
SEMAFOR (Reported by \cite{hartmann2017out})  & 60.01 & - && - & - \\
Hartmann et al. \cite{hartmann2017out}  & 62.51 & - && - & - \\
\textbf{PAFIBERT - Filtered by LUs} & \textbf{75.06} & \textbf{69.07} && \textbf{-} & \textbf{-} \\
\textbf{PAFIBERT - Filtered by Targets} & \textbf{75.01} & \textbf{69.07} && \textbf{-} & \textbf{-} \\
\bottomrule
\end{tabularx}
\end{table}

\begin{table}
\caption{Results without frame filtering}
\label{tab:result-nofil}
\renewcommand\arraystretch{1.4}
\begin{tabularx}{\textwidth}{X c c c c c}
\toprule
\multirow{2}{*}{\textbf{Model}} & 
\multicolumn{2}{c}{\textbf{FrameNet 1.5}} &&
\multicolumn{2}{c}{\textbf{FrameNet 1.7}} \\
\cmidrule{2-3}\cmidrule{5-6}
& \textbf{All} & \textbf{Ambiguous} && \textbf{All} & \textbf{Ambiguous} \\
\midrule
\multicolumn{6}{l}{\underline{\textit{Das's Test Set \cite{das2011semi}}}} \\
Hartmann et al. \cite{hartmann2017out} & 77.49 & - && - & - \\
Botschen et al. \cite{botschen-etal-2018-multimodal}  & 81.21 & 72.51 && - & - \\
\textbf{PAFIBERT - No Filter} & \textbf{89.57} & \textbf{81.86} && \textbf{88.97} & \textbf{81.83} \\
\midrule
\multicolumn{6}{l}{\underline{\textit{YAGS Test Set \cite{hartmann2017out}}}} \\
\textbf{PAFIBERT - No Filter} & \textbf{72.17} & \textbf{67.17} && \textbf{-} & \textbf{-} \\
\bottomrule
\end{tabularx}
\end{table}

Table \ref{tab:confused} shows the top five most frequently confused pairs of frames from the experiments. There is a substantial overlap in these confusing frames across models. In particular, our models seemed to have trouble distinguishing the following frames:

\begin{table}
\caption{Most frequently confused pairs of frames (for FrameNet 1.7)}
\label{tab:confused}
\renewcommand\arraystretch{1.6}
\begin{tabularx}{\textwidth}{l l C L}
\toprule
\textbf{Gold Frame} & \textbf{Predicted Frame} & \textbf{\# Misclassified Instances} & \textbf{Potentially Confusing LUs} \\
\midrule
\multicolumn{4}{l}{\underline{\textit{PAFIBERT - Filtered by Targets}}} \\
{[Possession]} & {[Have\_associated]} & 14 & have.v, have got.v \\
{[Possibility]} & {[Capability]} & 13 & can.v \\
{[Measure\_duration]} & {[Calendric\_unit]} & 12 & century.n, day.n, decade.n, fortnight.n, hour.n, millennium.n, minute.n, month.n, second.n, week.n, year.n \\
{[Political\_locales]} & {[Locale\_by\_use]} & 9 & country.n, city.n, village.n \\
{[Visiting]} & {[Arriving]} & 9 & visit.v, visit.n \\
\midrule
\multicolumn{4}{l}{\underline{\textit{PAFIBERT - No Filter}}} \\
{[Possibility]} & {[Capability]} & 13 & can.v \\
{[Possession]} & {[Have\_associated]} & 12 & have.v, have got.v \\
{[Capability]} & {[Possibility]} & 10 & can.v \\
{[Locative\_relation]} & {[Spatial\_co\_location]} & 9 & at.prep, here.adv, there.adv \\
{[Visiting]} & {[Arriving]} & 9 & visit.v, visit.n \\
\bottomrule
\end{tabularx}
\end{table}

\begin{itemize}
\item {[Possibility]} and {[Capability]}
\item {[Possession]} and {[Have\_associated]}
\end{itemize}

\begin{figure}
  \includegraphics[width=\linewidth]{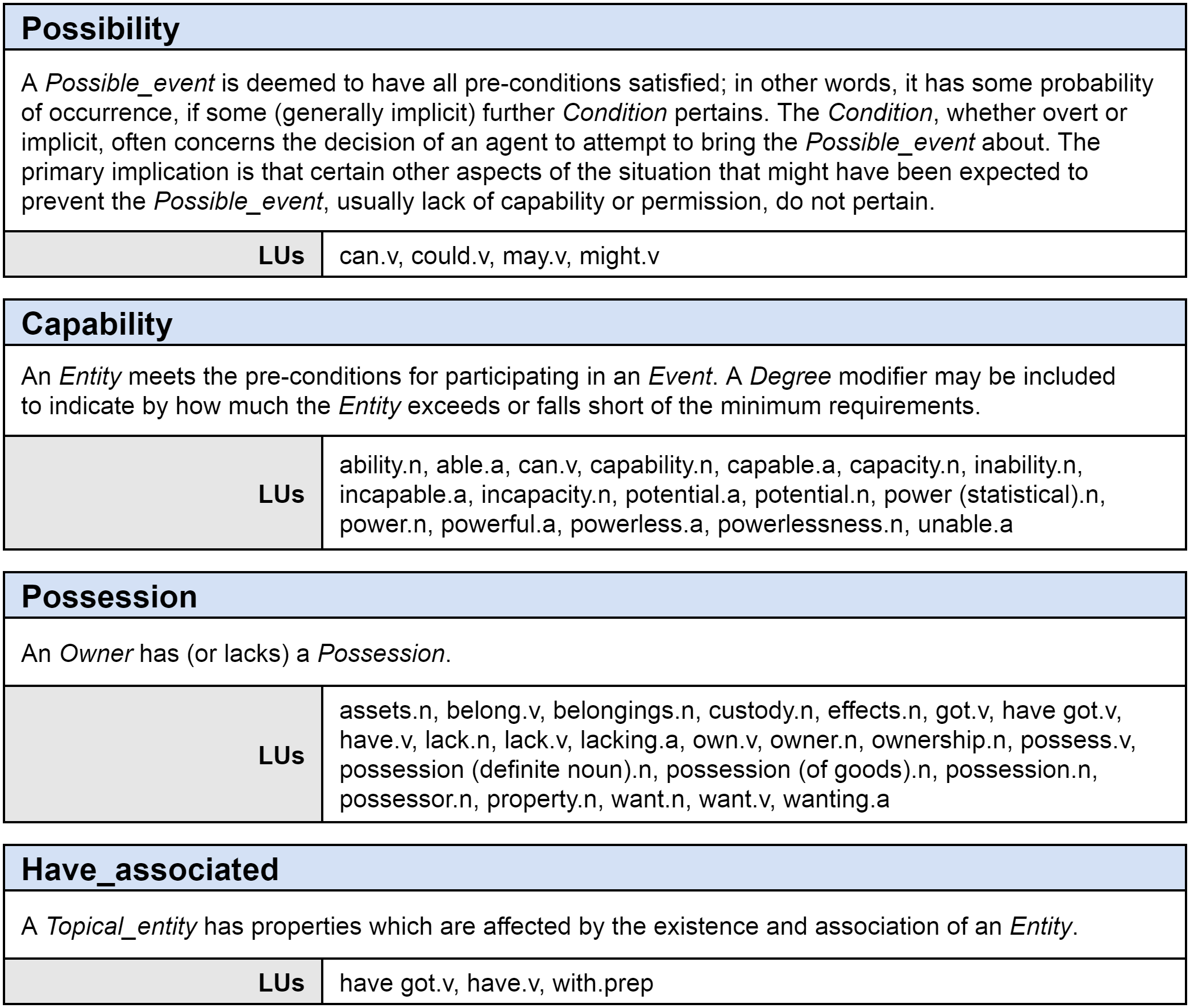}
  \caption{Simplified schematic representations for the  frequently confused frames}
  \label{fig:confused-frames}
\end{figure}

Some examples of the misclassified instances are shown in Table \ref{tab:misclassified}. Figure \ref{fig:confused-frames} gives the simplified schematic representations of the confused frames to aid the interpretation of the misclassified examples. From the examples, we found that recognizing the distinction between these confused senses is indeed quite tricky, and it requires a deep understanding of the underlying sentential contexts. Despite the difficulty in distinguishing these more nuanced senses, the overall improvements made by PAFIBERT are still encouraging. In particular, the performance gain for ambiguous targets is quite promising: On Das's test set, for example, PAFIBERT achieved absolute improvements of 4-5\% and 9\% with frame filtering (Table \ref{tab:result-fil}) and without frame filtering (Table \ref{tab:result-nofil}) respectively. As reported by \cite{hartmann2017out}, frame identification models usually suffer a drastic drop in performance when tested on out-of-domain data. The same trend was observed in our experiments for the YAGS test set. Nevertheless, PAFIBERT still outperformed existing methods by a large margin on this out-of-domain test set.

\begin{table}
\caption{Examples of instances misclassified by PAFIBERT with target-based frame filtering}
\label{tab:misclassified}
\renewcommand\arraystretch{1.6}
\begin{tabularx}{\textwidth}{L l l}
\toprule
\textbf{Test Sentence} & \textbf{Gold Frame} & \textbf{Predicted Frame} \\
\midrule
The people who \underline{can} benefit most directly from your generosity have no time to waste . & {[Possibility]} & {[Capability]} \\
You \underline{can} help them to know that feeling . & {[Capability]} & {[Possibility]} \\
Hong Kong is crowded - it \underline{has} one of the world 's greatest population densities . & {[Possession]} & {[Have\_associated]} \\
The casinos \underline{have} no admission charge and formal dress is optional , though long pants for men are required . & {[Have\_associated]} & {[Possession]} \\
\bottomrule
\end{tabularx}
\end{table}

Besides, in the more challenging setup that involved no frame filtering, the results obtained by PAFIBERT were on par with the prior state-of-the-art accuracies achieved with frame filtering. As will be demonstrated in the discussion below, PAFIBERT without frame filtering also has the advantages of generalizing well in other domains and making reasonable inferences for unseen targets.

\subsection{Qualitative Evaluation on Out-of-Domain Sample Sentences}

\begin{figure}
  \includegraphics[width=\linewidth]{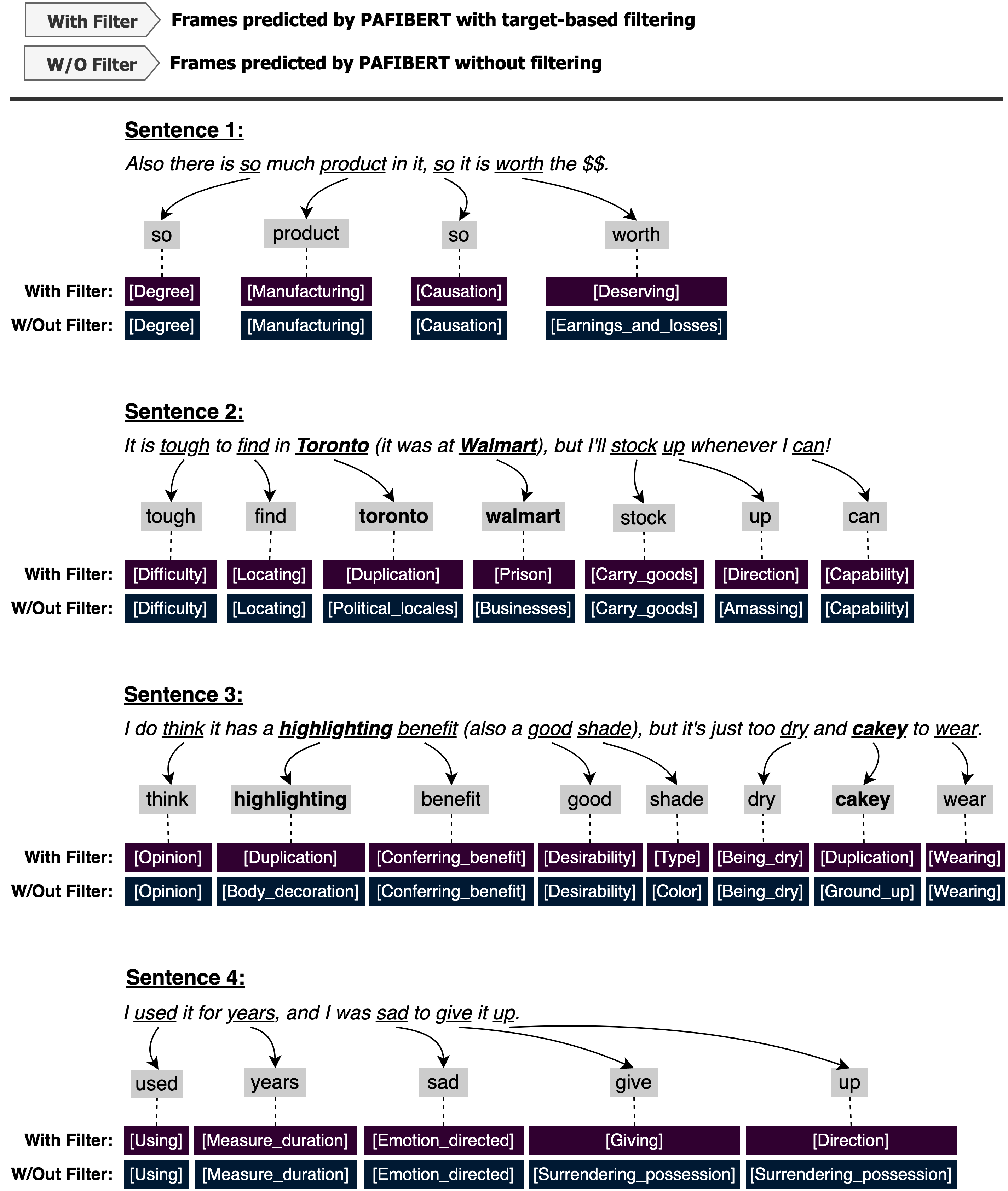}
  \caption{The sample sentences, targets, and predicted frames (based on FrameNet 1.7). Unseen targets are presented in boldface.}
  \label{fig:qualitative}
\end{figure}

In this section, we present interesting frame identification cases for several sample sentences selected from beauty product reviews. The predictions made by the two PAFIBERT variants---without frame filtering and with target-based filtering---are compared. Figure \ref{fig:qualitative} shows the sentences, targets, and predicted frames.

First and foremost, due to the way PAFIBERT was designed to utilize targets' positions for attention-based frame identification, both variants of PAFIBERT were able to handle the ambiguous targets that occur multiple times in the same sentence but refer to different frames with each occurrence. This scenario is exemplified by Sentence 1, in which the word `so' occurs twice. The first occurrence of `so' was linked to the [Degree] frame, whereas the second occurrence was recognized as referring to the [Causation] frame.

What stands out the most from this qualitative evaluation is that although PAFIBERT with frame filtering achieved higher accuracies on the benchmark test sets, it did not produce reasonable predictions for the unseen targets in the sample sentences. On the other hand, PAFIBERT without frame filtering seemed to handle the unseen targets more satisfactorily. It is generally agreed that inadequate coverage of LUs is a major shortcoming of FrameNet \cite{hartmann2017out, botschen-etal-2018-multimodal}. Previous studies have tried to address this issue in many ways: some extended FrameNet using other resources such as WordNet, VerbNet, and so forth \cite{johansson2007lth, shi2005putting, tonelli2009wikipedia}; others adopted statistical or graph-based strategies to generalize to unseen targets \cite{das2011semi, das2010semafor}. Nevertheless, unseen targets remain a major obstacle in applying frame-semantic parsing to real-world problems. From the examples in Figure \ref{fig:qualitative}, we observed that PAFIBERT without frame filtering selected suitable frames for most of the unseen targets, including `toronto' (Sentence 2), `walmart' (Sentence 2), `cakey' (Sentence 3), and `highlighting' (Sentence 3). Particularly interesting is the ability of the model to associate `toronto' with [Political\_locales] and `walmart' with [Businesses]. This behavior of the model resembles that of a named entity recognizer (e.g., \cite{finkel2005incorporating}), but it encompasses a broader range of categories in comparison to the latter. 

Also, since meanings of words are sometimes domain-specific, another attractive property of PAFIBERT without frame filtering is its ability to identify frames in accordance with the domain. For instance, associating the word `highlighting' with [Body\_decoration] (Sentence 3) is appropriate in the beauty domain because a highlighter is a cosmetic product applied to some regions of the face to reflect light and create contours. Another domain-aware identification is exemplified by the word `shade' (Sentence 3). Although `shade.n' is listed under the [Type] frame in FrameNet, it was tagged as [Color] by PAFIBERT without frame filtering. Indeed, in this specific context, [Color] seems to be a better match for `shade.'

Furthermore, PAFIBERT without frame filtering worked better for non-contiguous multi-word LUs, i.e., LUs that contain insertions. In Sentence 4, for example, when the constituents of the LU `give up.v' are separated by the word `it,' PAFIBERT with frame filtering failed to associate `up' with [Surrendering\_possession] because this frame is not a candidate frame for `up.' However, without frame filtering and the constraints it enforces, PAFIBERT was able to recognize the word `up' as part of the phrasal verb `give up.'

So far, PAFIBERT without frame filtering is in the lead in the above comparison of model generalizability and ability to handle unseen targets. However, there is a disadvantage in performing frame identification without frame filtering. Specifically, since a no-filtering model is not bounded by candidate frames, it relies solely on sentential contexts to make its predictions. This renders the model highly sensitive to contextual variations. In other words, the model might be more prone to identification errors when the context it encountered in a sentence differed significantly from the typical contexts seen in the training data. In contrast, PAFIBERT with frame filtering can take full advantage of the frame-LU relations defined in FrameNet to avoid straying from the correct frames, thus reducing the risk of being misled by unfamiliar contexts. For instance, with frame filtering, the target `worth' in Sentence 1 was treated as unambiguous because FrameNet only returned [Deserving] as the candidate frame for this target. However, due to its sole reliance on sentential contexts, PAFIBERT without frame filtering chose [Earnings\_and\_losses] for this target. Although the predicted frame is not too far-fetched, it does not seem to be a better match than [Deserving]. This issue creates a dilemma because while frame filtering can reduce the errors caused by contextual variations, enforcing candidate frames would reduce the generalizability of frame identification models. Future research should attempt to find a hybrid solution that biases frame predictions toward the candidate frames specified by FrameNet while maintaining the ability to deal with unseen targets satisfactorily and to choose other frames over candidate frames if deemed more suitable.

\section{Conclusions}

This paper presented a neural network architecture called Positional Attention-based Frame Identification with BERT (PAFIBERT) as a solution to frame identification, which is an indispensable subtask in frame-semantic parsing. PAFIBERT inherited the language representation power of BERT and combined it with a position-based attention mechanism to capture target-specific contextual information. The comparison with existing models revealed that PAFIBERT was empirically powerful and superior to other models. Under various experimental settings, PAFIBERT showed significant improvements for both in-domain and out-of-domain test sets. Two variants of PAFIBERT were created and compared in this study, and each variant has its own strengths and weaknesses. Future research may work toward finding a solution that combines the best of both worlds. Besides, through the analysis of the most frequently confused pairs of frames, we found that PAFIBERT still failed to recognize some subtle differences that required a more profound understanding of sentence contexts and semantics. Considerably more work will need to be done to find out how far the rich semantic representation of pre-trained language models can take us in dealing with this kind of nuanced sense distinction. Since the top frequently confused frames caused a substantial loss in model accuracies, another possible progression of this work is to consider a divide-and-conquer strategy that places additional emphasis on solving the more challenging cases.

\section*{Acknowledgements}
The research leading to the results in this study has received funding from Nanyang Technological University, Singapore, under grant agreement M4082112.

\bibliographystyle{unsrt}
%\bibliography{references}  %%% Remove comment to use the external .bib file (using bibtex).
%%% and comment out the ``thebibliography'' section.

\end{document}